\documentclass[letterpaper, 10 pt, conference]{ieeeconf}  

\IEEEoverridecommandlockouts                              

\usepackage{graphicx}
\usepackage{amsmath}
\usepackage{amssymb}
\usepackage{adjustbox}
\usepackage{algorithm}
\usepackage{booktabs}
\usepackage{diagbox}
\usepackage[noend]{algpseudocode}
\usepackage{caption}
\usepackage{cite}
\usepackage[colorlinks]{hyperref}
\usepackage{balance}

\title{\LARGE \bf
RIC: Rotate-Inpaint-Complete for Generalizable Scene Reconstruction
}

\author{Isaac Kasahara, Shubham Agrawal, Selim Engin, Nikhil Chavan-Dafle, Shuran Song, and Volkan Isler
\\
Samsung AI Center, New York
}

\newcommand{\dalle}{DALL·E 2}
\newcommand{\ours}{RIC}

\begin{document}

\maketitle
\thispagestyle{empty}
\pagestyle{empty}

\begin{abstract}

General scene reconstruction refers to the task of estimating the full 3D geometry and texture of a scene containing previously unseen objects. In many practical applications such as AR/VR, autonomous navigation, and robotics, only a single view of the scene may be available, making the scene reconstruction task challenging. 
In this paper, we present a method for scene reconstruction by structurally breaking the problem into two steps: rendering novel views via inpainting and 2D to 3D scene lifting. Specifically, we leverage the generalization capability of large visual language models (\dalle{}) to inpaint the missing areas of scene color images rendered from different views. Next, we lift these inpainted images to 3D by predicting normals of the inpainted image and solving for the missing depth values.  By predicting for normals instead of depth directly, our method allows for robustness to changes in depth distributions and scale.
With rigorous quantitative evaluation, we show that our method outperforms multiple baselines while providing generalization to novel objects and scenes.

\end{abstract}

\section{INTRODUCTION}


The understanding of 3D scene geometry is essential for many down-stream applications. In robotics, it allows for accurate manipulation and motion planning considering the surrounding environment. In the field of augmented reality, it allows for better mapping and rendering to bridge the virtual world to the real world. With smartphones and robots that are equipped with high quality depth sensors, the task of 3D scene reconstruction is becoming feasible in such domains. 

\begin{figure}[ht!]
    \centering
    \includegraphics[width=0.9\columnwidth]{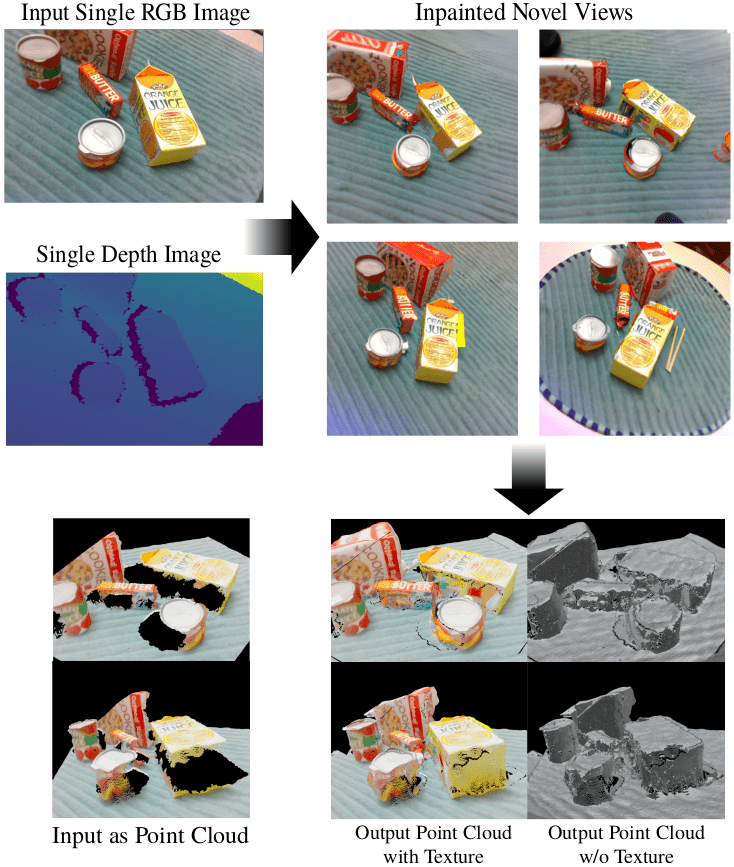}
    \caption{\ours{} inputs a single RGB-D image and generates complete 3D scene reconstruction with texture (bottom-left). RIC first \textbf{R}otates the input RGB-D image to a novel viewpoint, \textbf{I}npaints the missing regions using generalizable visual language models, and finally \textbf{C}ompletes the depth via normal prediction and optimization. For comparison, input as a point cloud is also shown on the bottom-left.}
    \label{fig:figure-1}
    \vspace{-8mm}
\end{figure}

These depth sensors allow for accurate reconstruction of the observed parts of the scene. However, to reconstruct the unseen parts, we must use prior information conditioned on the observed information. The missing information in the input image combined with the diversity in shapes, sizes, and depth distribution of the household objects presents a major challenge for scene reconstruction in-the-wild. 
In this paper, we study this problem in a general setting, where the goal is to reconstruct a complex scene with multiple novel objects, given only one RGB-D image of the scene. 

We present our method Rotate-Inpaint-Complete (\ours{}), which predicts both the 3D geometry and the texture of the unseen parts of the scene in the input image by leveraging the inpainting capabilities of large visual-language models.
Given an RGB-D image of a scene, first we generate novel views (RGB and depth images) by rotating and then projecting the input scene. Then we use a surface-aware masking method to select regions in the image to allow us to inpaint utilizing the powerful 2D inpainting capabilities of \dalle{}~\cite{ramesh2022hierarchical} for exposing the potential object geometry not visible in the input image. 
Finally, we optimize the depth images using the input depth values, as well as the occlusion boundaries and normals estimated from the inpainted images. These inpainted and completed novel RGB-D views provide the reconstructed scene geometry as a fused point cloud with associated textures.
To mitigate the object hallucination and spatial inconsistency of predictions from \dalle{}, we use a consistency filtering method to enforce consistency across viewpoints which plays a crucial role for generalizable, yet accurate and robust scene reconstruction.



\begin{figure*}[ht!]
    \centering
    \includegraphics[width=0.9\linewidth]{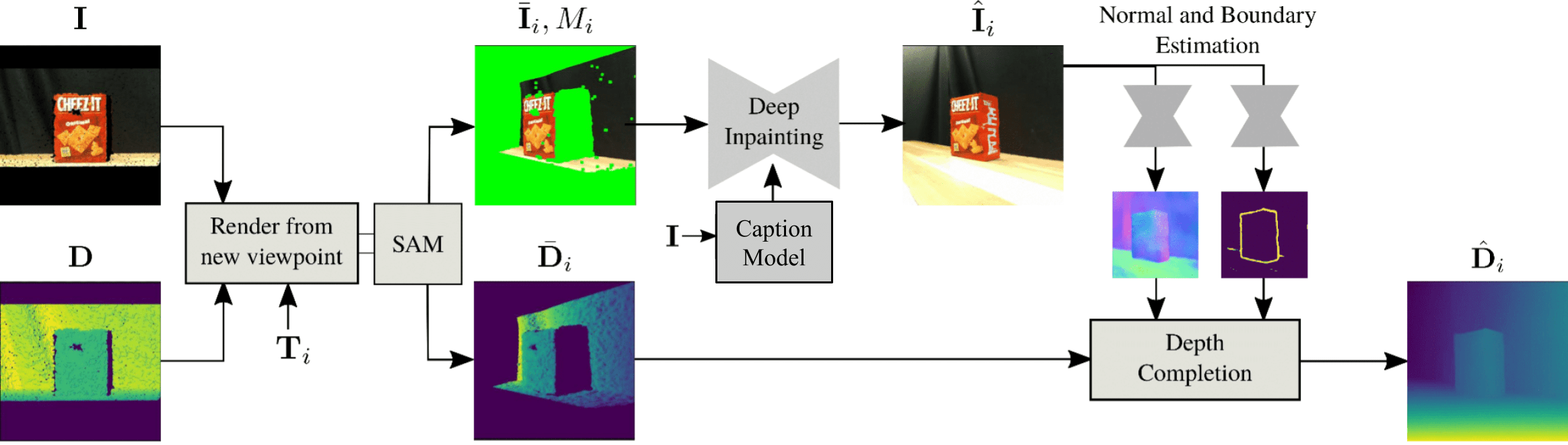}
    \caption{\textbf{Method Overview:} \ours{} takes as input an RGB-D image and starts by rendering incomplete RGB-D images $\bar{\mathbf{I}}_i$ and $\bar{\mathbf{D}}_i$ from a new viewpoint $\mathbf{T}_i$. The missing RGB values of $\bar{\mathbf{I}}_i$ are inpainted using a diffusion-based VLM given a generated prompt, such as ``a photo of household objects on a table", where the pixels to be inpainted are determined by our Surface-Aware Masking (SAM) technique. The inpainted image is used to predict surface normals and occlusion boundaries at the new viewpoint $\mathbf{T}_i$, which are then used for completing the missing depth values along with the incomplete depth image $\bar{\mathbf{D}}_i$. After repeating this process for $V$ viewpoints, the final output of \ours{} is a merge of deprojected depth predictions.}
    \label{fig:method-overview}
    \vspace{-5mm}
\end{figure*}

In short, the contributions of this paper can be summarized as follows. \textit{i)} We present an integrated approach for scene completion of unseen objects under occlusion and clutter, by solving the problem through novel view inpainting and 2D to 3D scene lifting. \textit{ii)} We develop a method for selectively inpainting regions in the novel views of the input scene that enables synthesis of consistent 2D geometry. \textit{iii)} We train a 2D to 3D lifting method on the YCB-V~\cite{xiang2018posecnn} dataset and demonstrate the generalization capability to cluttered scenes containing novel household objects and categories.

\section{RELATED WORK}

\textbf{Scene Reconstruction:}
While single-object reconstruction is a well-studied problem, full-scene reconstruction is explored in limited settings. Previously works in scene reconstruction were focused either on room scale~\cite{song2017semantic, dai2018scancomplete} or in autonomous driving settings~\cite{cheng2021s3cnet, rist2021semantic, cao2022monoscene} where the scene geometries are usually more structured. In this work, we focus on an object-level scale, specifically tabletop environments, where objects can be in cluttered configurations. While methods like ~\cite{gkioxari2019mesh, popov2020corenet, irshad2022centersnap} 
show an accurate reconstruction of objects at the scene level, they do not generalize to novel category objects. Recently,~\cite{wu2023multiview} introduced a method for reconstructing 3D geometries of objects and scenes of unseen categories, and demonstrated generalization capability to objects in-the-wild. However, different from our setting, they mostly focus on isolated objects and scenes with little to no clutter. In contrast, our method can reconstruct geometries and textures of complex scenes with objects from novel categories under heavy occlusions, as we show in our experiments.

\textbf{Inpainting:}
While traditional inpainting methods made use of hand-crafted image priors to fill small gaps for tasks like image restoration \cite{elharrouss2020image}, deep generative methods like Generative Adversarial Networks (GAN) ~\cite{goodfellow2020generative} have shown remarkable success for tasks like image denoising~\cite{chen2018image}, super-resolution~\cite{ledig2017photo}, and inpainting~\cite{pathak2016context, iizuka2017globally, zhao2021large}. However, GAN models are known for potentially unstable training for large datasets \cite{weng2021diffusion}. More recently, resulting from the growth of visual language diffusion models, which can be efficiently trained on internet-scale datasets, inpainting through image diffusion has shown great generalization capabilities to many different objects and scenes~\cite{ramesh2022hierarchical}.
In this paper, we develop a process to use a visual language diffusion model for inpainting cluttered scenes involving heavy occlusions.

\textbf{Text-to-3D Synthesis: }
Recent papers such as~\cite{jain2022zero, poole2022dreamfusion, lin2022magic3d} have demonstrated the ability to generate 3D models of very diverse objects from merely a text description. Despite the realistic appearance, these generated objects are not grounded to any real-world geometry.
To overcome this limitation, ~\cite{xu2022neurallift, melas2023realfusion} extended these methods to reconstruct based on a ground truth reference image. These papers demonstrate high accuracy on individual objects, but do not demonstrate the ability to reconstruct multiple objects in cluttered scenes. Moreover, their runtime is a concern, as optimizing neural radiance fields ~\cite{mildenhall2021nerf} can take up to an hour. ~\cite{nichol2022point} attempts to produce faster results by optimization without NeRF but are still limited to single object reconstruction and do not directly generalize to our setting. 


\section{METHOD}



In this section, we present our method Rotate-Inpaint-Complete, or \ours{}, for generalizable reconstruction of a 3D scene containing multiple objects, given a single RGB-D image of the scene.

\ours{} takes in as input an RGB-D image \(\mathcal{I} = (\mathbf{I}, \mathbf{D}) \in \mathbb{R} ^{ H \times W \times 4} \) and outputs a color point cloud \(S \in \mathbb{R} ^{ N \times (3 + 3)} \), where $N$ is the number of predicted points in the scene.
Our method consists of three main components: 1) An inpainting step that takes in an RGB-D image \(\mathcal{I}\) and outputs an inpainted RGB image \(\hat{\mathbf{I}}_i\) from a novel viewpoint $\mathbf{T}_i \in SE(3)$. 2) A depth completion component that takes in the inpainted RGB image \(\hat{\mathbf{I}}_i\) as well as an incomplete depth image \(\bar{\mathbf{D}}_i \) rendered from the viewpoint $\mathbf{T}_i$, and outputs a completed depth \(\hat{\mathbf{D}}_i\) at that viewpoint. 3) A viewpoint selection and consistency filtering method that utilizes the above two components to generate completed RGB-D images at rotated novel views and uses them to reconstruct the scene. We explain each of these components in detail next.

\subsection{Inpainting}

This section describes the inpainting process, as well as the intermediate steps taken before and after to go from the input image \(\mathbf{I}\) to \(\hat{\mathbf{I}}_i\) at a novel viewpoint $\mathbf{T}_i$.

\subsubsection{Rotate and Project RGB-D Image}
Given an RGB-D image of a scene and the camera intrinsics, we deproject the image into a point cloud in the camera frame. This point cloud is then projected onto a novel viewpoint $\mathbf{T}_i$ and the resulting image is masked using our Surface-Aware Masking method (SAM), which we describe in detail in the following section. The projection from this new viewpoint creates a new RGB-D image $\bar{\mathbf{I}}_i$ with missing RGB and depth information as seen in Figure~\ref{fig:method-overview}. Small holes of the missing RGB values are filled with a naive inpainting algorithm~\cite{telea2004image} by inpainting pixels that are covered after a morphological closing operation of kernel size 5 is applied to the mask. The larger missing areas are left for the deep inpainting module.

\subsubsection{Surface-Aware Masking}
\label{sec:bg-mask}

\begin{figure}[ht!]
    \centering
    \includegraphics[width=0.95\columnwidth]{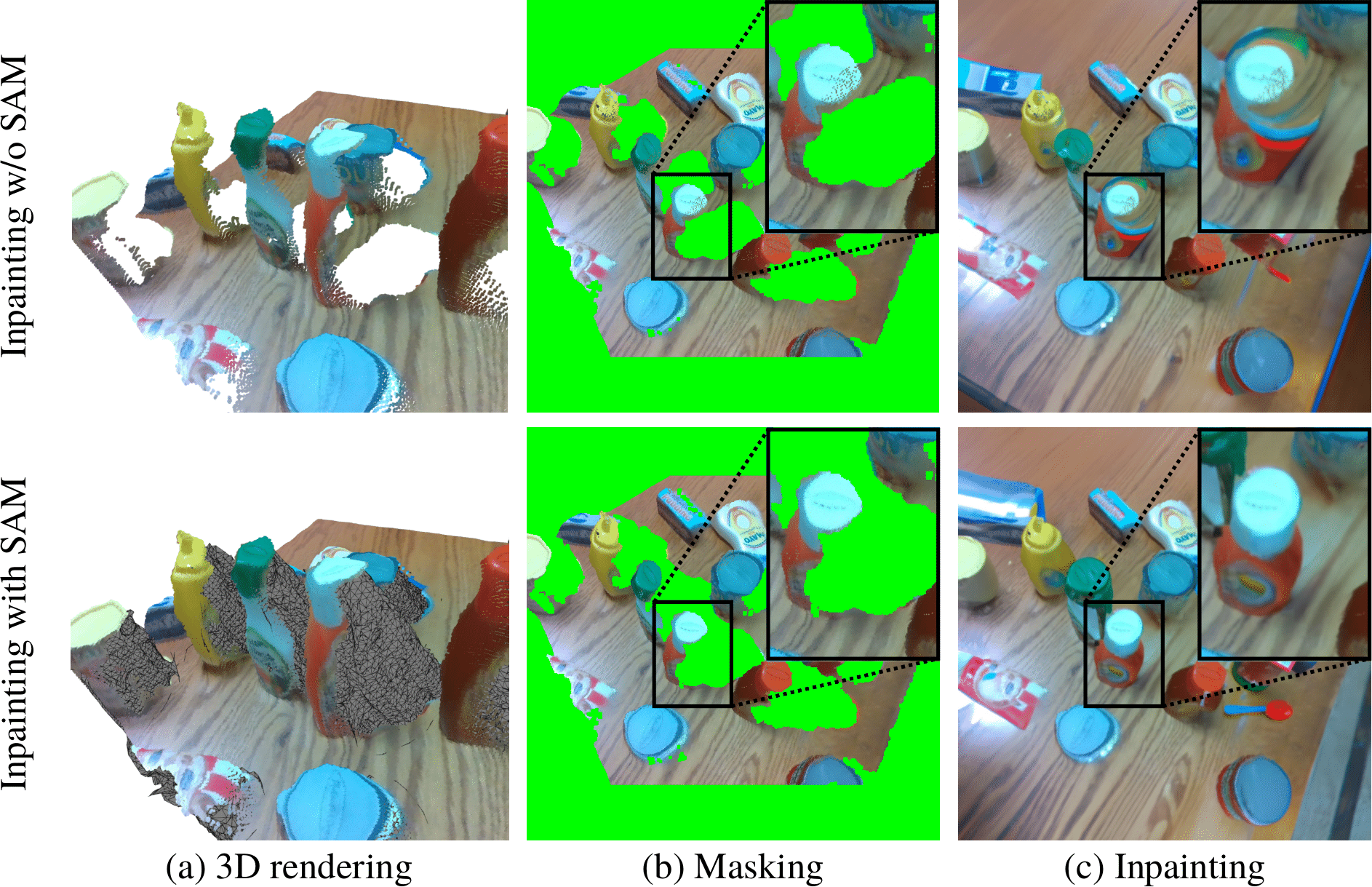}
    \caption{Surface-Aware Masking (SAM) is a necessary step to obtain realistic inpaintings.
    Naively rotating the input point cloud moves the background pixels next to the foreground object pixels (b-top) which results in poor inpainting (c-top). Using SAM, we correctly mask out the background pixels which results in good inpainting results (b-bottom).}
    \label{fig:sam_fig}
    \vspace{-3mm}
\end{figure}

In order for inpainting to work properly, a mask covering the areas to inpaint needs to be generated. After projecting to the new camera frame, any 3D space possible to be reconstructed needs to be represented as an inpainting mask in the 2D image. This issue can be seen in Figure~\ref{fig:sam_fig} as the table takes up pixels we may want to fill in with the bottle. To solve this problem, a 3D frustum is generated from the original camera and depth image. For every pixel in the original camera frame, a ray is cast from the camera through each point in the projected point cloud from \(\mathcal{I}\). Once the ray has passed through its respective point, it is used to generate a list of points along the ray from that depth onward with $m$ points of equal spacing $c$. This is done for every ray, and from this process results a point cloud covering the potential space that the 3D scene could possibly fill. This point cloud is then converted to a mesh, and when the point cloud from the RGB-D image \(\mathcal{I}\) is rotated to novel views, the mesh is rotated with it. Finally, when projecting back to the camera frame after rotation, points that are occluded by the mesh are discarded. Any blank pixels are then used as the 2D inpainting mask to be filled when passed to the inpainting step. This procedure of generating the final image and mask is detailed in our technical report (see Appendix) and its outputs are shown in Figure~\ref{fig:method-overview}, with the green pixels representing the inpainting mask.

\subsubsection{Diffusion-based Inpainting}
\label{sec:diff-inpaint}
Once these preprocessing steps have been completed, we pass the processed image and a mask of areas to be filled in to the inpainting algorithm. We use \dalle{}\cite{ramesh2022hierarchical} for image inpainting since it demonstrates the ability to produce the most realistic results. This model takes in the incomplete image \(\bar{\mathbf{I}}_i\), the mask generated in the previous step \(M\), and an input prompt \(P\) that describes the context of the image in words. 
For prompt, we pass the RGB image $\mathbf{I}$ to a deep captioning model ~\cite{li2022blip} and prefix the generated caption with \textit{``A photo of''}. 
We also explore using a more specific and generic prompt in our ablation experiments (Table~\ref{tab:prompt}).
The output from this inpainting method is an image  \(\hat{\mathbf{I}}_i\) that now contains estimated areas from the diffusion model. Figure~\ref{fig:method-overview} shows an example before and after inpainting with \dalle{}.


\subsection{Depth Completion}
We use a method proposed in~\cite{zhang2018deepdepth} for generating a complete depth image \(\hat{\mathbf{D}}_i\) from an incomplete depth image \(\bar{\mathbf{D}}_i\) and its corresponding RGB image. This method estimates the normals and occlusion boundaries from the RGB image, and optimizes for the complete depth by utilizing the estimated normals, occlusion boundaries, and incomplete depth.

\subsubsection{Normals and Occlusion Boundaries Prediction}
In order to obtain estimations for the normals and occlusion boundaries, we train Deeplabv3+ with DRN-D-54 in the same manner as in~\cite{sajjan2020clear}. 
The ground truth normals and occlusion boundaries are obtained using the depth images from the YCB-V training dataset~\cite{xiang2018posecnn}, the YCB-V synthetic dataset \cite{denninger2020blenderproc, hodan2020bop}, and the HomebrewedDB synthetic dataset~\cite{kaskman2019homebreweddb}. 

\subsubsection{Optimize for Depth}
Given the incomplete depth, the estimated normals from the image, and estimated occlusion boundaries, we solve for the completed depth. The main idea behind this method in~\cite{zhang2018deepdepth} is that the areas with missing depth can be computed by tracing along the estimated normals from areas of known depth with the occlusion boundaries acting as barriers where normals should not be traced across. Formally we solve a system of equations to minimize an error \(E\), where \(E\) is defined as \(E=\lambda_DE_D + \lambda_SE_S + \lambda_NE_NB\). Here, \(E_D\) is the distance between the ground truth and estimated depth, \(E_S\) influences nearby pixels to have similar depths, \(E_N\) measures the consistency of estimated depth and estimated normal values, and \(B\) weights the normal values based on the probability that it is a boundary. We use the same \( \lambda_D,\lambda_S,\lambda_N \) values as in~\cite{sajjan2020clear}. 

\subsection{Scene Completion}
This section describes the complete process we follow to reconstruct a 3D scene from a single RGB-D image.


\subsubsection{Viewpoint Selection} \label{sec:pcl_rotation}

For diffusion-based inpainting, ``known" pixels, i.e., the non-masked areas, guide the prediction of the unknown masked areas. We refer to the known pixels as context pixels and define the context ratio \(C\) for any given image as \(C = (\# context~pixels)\: /\: (\# all\: pixels)\). This ratio gives us some indication about how accurately the inpainting model will be able to fill in the missing areas. With a low \(C\), many areas are unknown and inpainting will struggle, and with a high \(C\) inpainting will do well but only fill in minimal information. An example of different context ratio values can be seen in Figure~\ref{fig:figure-3}.

We then design our viewpoint selection process to search for a context ratio that will allow for accurate inpainting. To do this, we define a sphere with a center as the mean of the input point cloud, and the radius as the distance between the center and the initial camera location. Then from the starting viewing angle, we rotate in various directions along this sphere away from the starting position. At each step in this rotation, we project the input point cloud onto the new camera location. Using this newly projected image, we compute the context \(C\) of the image. If the \(C\) is closest to our chosen context threshold \(C^*\), we use that viewpoint $\mathbf{T}_i$ as next to inpaint. We repeat this process for \(V\) evenly spaced directions we traverse along the sphere as visualized in Figure~\ref{fig:figure-3}, where both \(C^*\) and \(V\) are chosen using the experiment described in Section~\ref{sec:implementation}.



\subsubsection{Enforcing Consistency Across Viewpoints}\label{sec:consistency}
{The final step in our method involves combining these generated viewpoints while enforcing consistency across them. One drawback of utilizing \dalle{} for inpainting real objects, is its inconsistent completion of objects as well as the hallucination of objects that are not originally in the scene. To combat this issue, we filter for consistent predictions across viewpoints. The final prediction is achieved by first deprojecting the RGB-D images from each viewpoint \(\mathbf{T}_i\) back into the original camera frame as point clouds. We then apply the following \textit{consistency rule} across all the generated points: If a predicted point from one viewpoint has a predicted point within a 1cm radius from at least two other viewpoints we keep that point, otherwise we remove that point from our final prediction. This rule allows us to keep points that only multiple viewpoints predict. We then combine all filtered points to obtain our final output point cloud of the completed scene \({S}\), which contains more accurate geometry and color than without filtering as seen in Table~\ref{tab:ablations} and Figure~\ref{fig:figure-1}.

\begin{figure}[ht!]
    \centering
    \includegraphics[width=0.8\columnwidth]{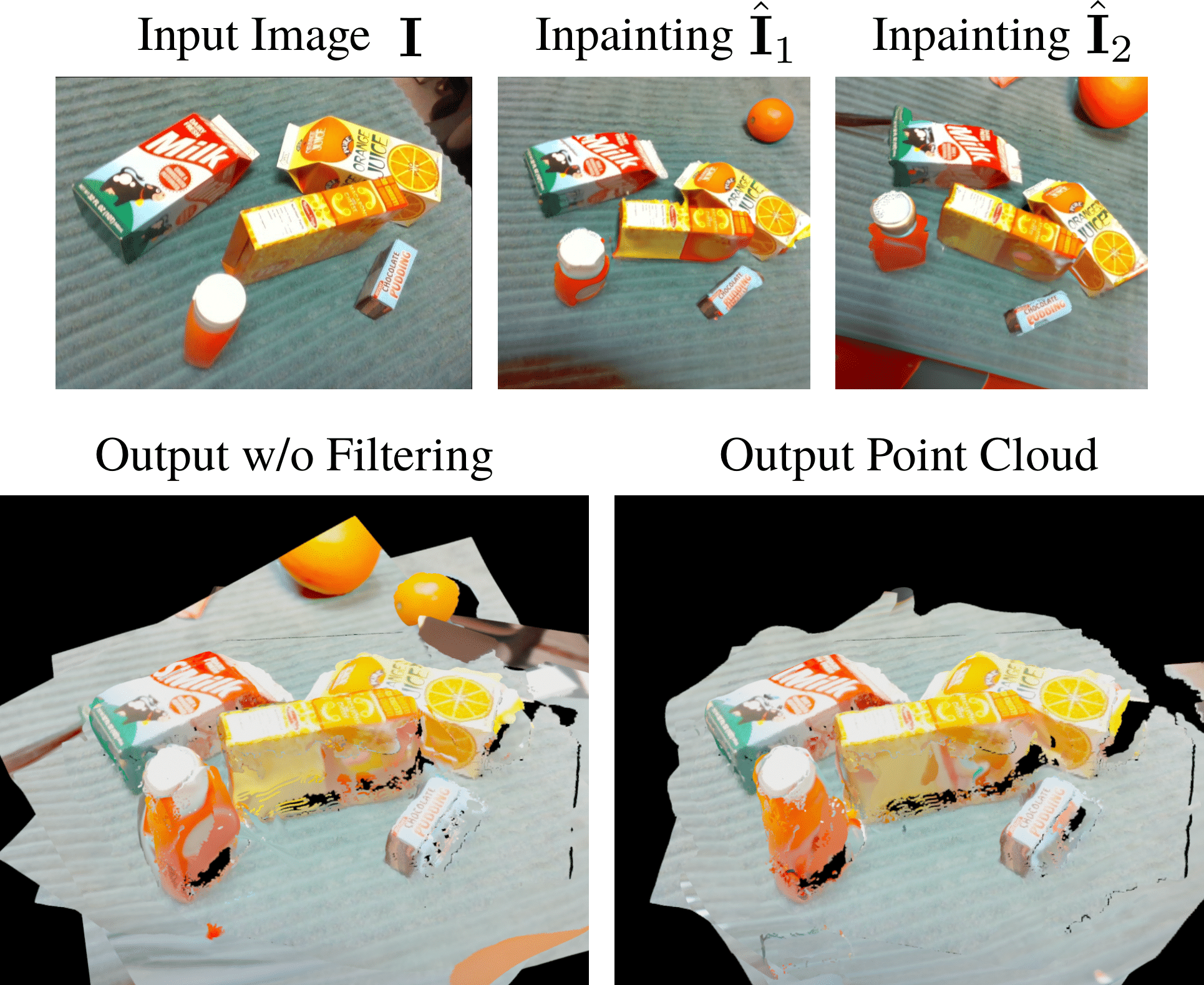}
    \caption{The consistency filtering step is used to remove the hallucinated objects in 3D, e.g., the oranges in the top right of the images get filtered out. For simplicity we visualize our consistency filtering step for only two viewpoints, while we filter using all viewpoints in our main method.}
    \label{fig:figure-1}
    \vspace{-3mm}
\end{figure}

\begin{figure}[ht!]
    \centering
    \includegraphics[scale=0.33]{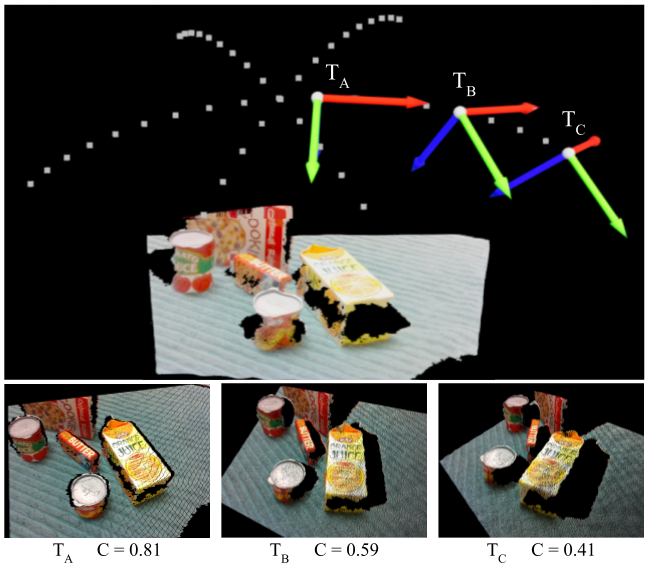}
    \caption{\ours{} samples viewpoints on evenly spaced directions along the viewing sphere (white dots). Several  viewpoints (top) as well as their corresponding rendered images are shown together with the context ratio of each image (bottom).}
    \label{fig:figure-3}
    \vspace{-5mm}
\end{figure}

\section{EXPERIMENTS}

In this section, we evaluate the performance of \ours{} for single view RGB-D scene reconstruction task. 
 We also report ablation studies for understanding the dependence of our method on (1) prompt specificity, (2) inpainting model, and (3) consistency filtering.

\textbf{Implementation Details: }
\label{sec:implementation}
The inpainting step of our algorithm is based on OpenAI's \dalle{} API.
For our implementation of SAM, we choose a spacing value $c$ of 0.01 meters with $m$ = 100 points for generating our rays. For choosing the number of viewpoints/viewpoint directions \(V\) as well as the context ratio \(C^*\) described in our method section, we perform a parameter search using 4 held out validation scenes from the YCB-V test set. We test using 6, 8, 10, and 12 viewpoints as well as a value of 0.3, 0.4, 0.5, 0.6, and 0.7 for our context threshold. We found that 10 views and 0.4 as a context threshold gave us the best accuracy on the validation set. 12 views and 0.4 as a context threshold performed similarly, but in the interest of runtime we use 10 for the final method. 
The full parameter grid search are provided in our appendix. For our module that enforces consistency between synthesized views, we choose a threshold of 0.01 meters when computing the intersection between points in the viewpoints point clouds.



\textbf{Datasets:}
We trained our depth completion model using the YCB-V training dataset \cite{xiang2018posecnn}. For testing, we test on $8$ unseen scenes from the YCB-V test set, and select $5$ RGB-D images from each of the scenes, i.e., we test on a total of $40$ RGB-D images in total. For ground truth point clouds, we deproject the RGB-D frames of the scene and concatenate them together. We also place the ground truth meshes in the scene for the objects and convert those to point clouds before concatenating them as well. Finally, we crop this point cloud around the ground truth meshes with a 10cm buffer as the RGB-D frames may contain floors and walls far away that we are not interested in reconstructing. This creates our final ground truth point cloud covering the majority of the scene with full geometry of the objects in the scene.

To demonstrate our model's capabilities of generalizing to unseen objects and to entirely new datasets, we also compare our method on the HOPE dataset \cite{tyree2022hope}. HOPE test set only contains individual RGB-D images and is unusable for generating full scene point cloud. Instead, we use HOPE training dataset for evaluation, as the train set contains RGB-D video and cluttered tabletop scenes with novel objects. The dataset has 10 scenes, and we again sample 5 frames per scene for a total of 50 RGB-D test images. Ground truth point clouds are obtained similarly to the YCB-V dataset.



\textbf{Baselines:}
We compare \ours{} against four baselines: Convolutional Occupancy Networks (CON) ~\cite{peng2020convolutional} is a 3D scene reconstruction method that inputs a sparse point cloud. We use their pre-trained model for \textit{Synthetic Indoor Scene dataset} where similar to our YCB-V and HOPE datasets, they place multiple ShapeNet ~\cite{shapenet2015} objects in indoor scenes. CoReNet ~\cite{popov2020corenet} is a multi-object shape estimator that inputs an RGB image and estimates a mesh. We compare against CoReNet's pre-trained model qualitatively since its predictions lack scale information. ShellNet ~\cite{chavan2022simultaneous} is trained for single object reconstruction. Given a scene depth image and object instance mask, ShellNet produces reconstruction for the object instance. We re-implemented ShellNet's architecture and trained it with Mask R-CNN~\cite{DBLP:journals/corr/HeGDG17} as the segmentation network on YCB-V dataset.
Finally, we compare against CenterSnap ~\cite{irshad2022centersnap}, a multi-object point cloud prediction method. CenterSnap inputs an RGB-D image and predicts point clouds for each object in the scene. Similar to CenterSnap's original training, we first train it on YCB-V synthetic dataset \cite{denninger2020blenderproc, hodan2020bop}, then fine-tune it on the YCB-V real training dataset \cite{xiang2018posecnn}. Since CenterSnap and ShellNet only predict the point clouds for objects and not the rest of the scene, for a fair evaluation, we concatenate their outputs with deprojected point cloud from the input RGB-D image.


\begin{figure}[ht!]
    \centering
    \includegraphics[width=0.9\columnwidth]{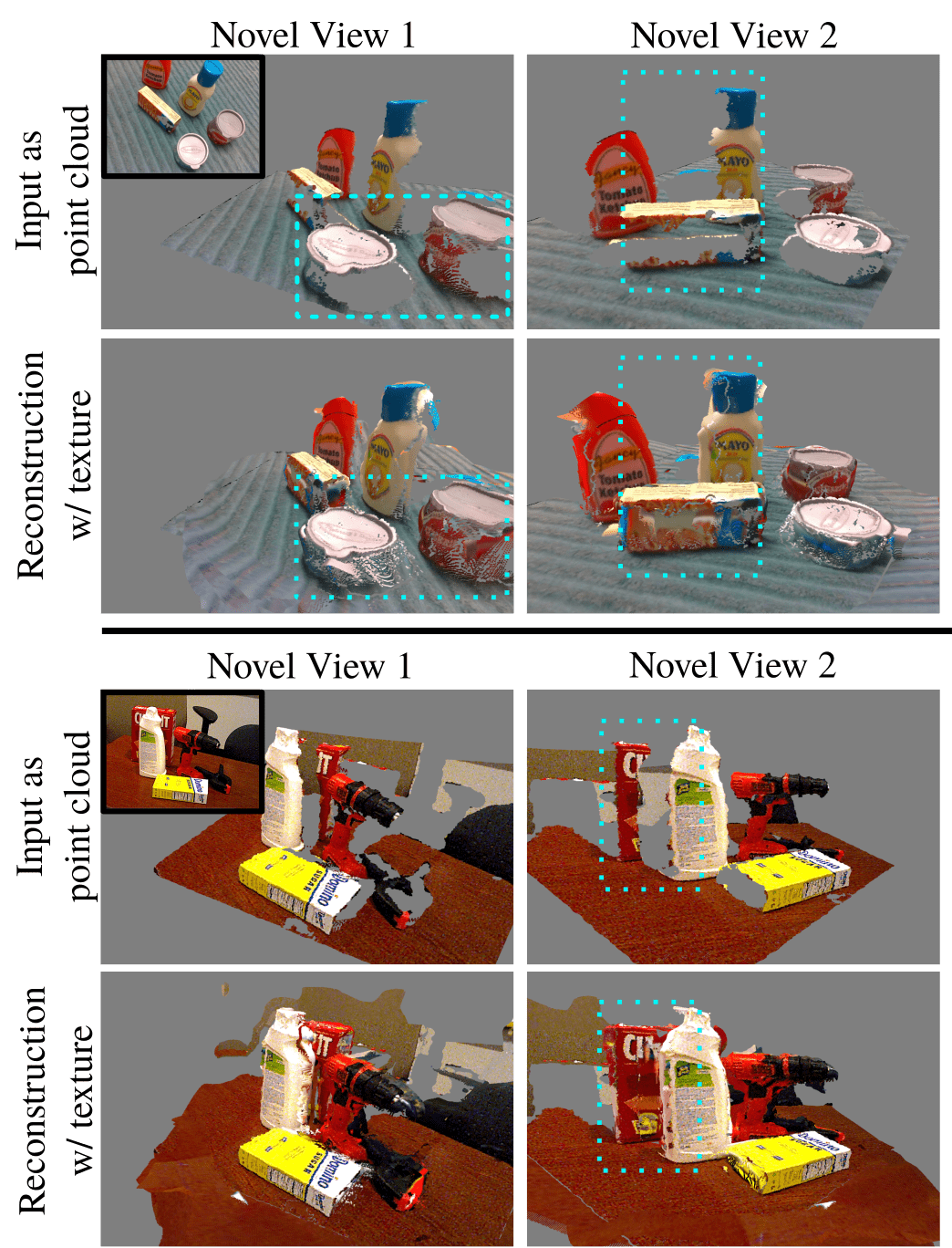}
    \caption{\textbf{Qualitative Results:} We show our scene completion results given a single RGB-D image, as color point clouds from two viewpoints. Top two rows are from the HOPE dataset~\cite{lin2021fusion}, and the bottom two are from YCB-V~\cite{xiang2018posecnn}.}
    \label{fig:qualitative-geometry}
    \vspace{-5mm}
\end{figure}


\begin{table}[t]
\begin{adjustbox}{width=\columnwidth,center}
\centering
\begin{tabular}{l | c c c c c}
\toprule
 \textbf{Method} & IoU $\uparrow$ & F-Score $\uparrow$ & CD$(S^*, S)$ $\downarrow$ & CD$(S, S^*)$ $\downarrow$ & CD $\downarrow$ \\
\toprule
\multicolumn{6}{c}{YCB-V~\cite{xiang2018posecnn}} \\
\midrule
 CON~\cite{peng2020convolutional} & 0.087 & 0.354 & 0.036 & 0.014 & 0.050 \\ 
 ShellNet~\cite{chavan2022simultaneous} & 0.224 & 0.607 & 0.019 & 0.012 & 0.031 \\
 CenterSnap~\cite{irshad2022centersnap} & 0.225 & 0.622 & 0.019 & \textbf{0.009} & \textbf{0.028} \\
 \ours{} (Ours) & \textbf{0.294} & \textbf{0.661} & \textbf{0.018} & 0.010 & \textbf{0.028} \\
\midrule
\multicolumn{6}{c}{HOPE~\cite{lin2021fusion}} \\
\midrule
CON~\cite{peng2020convolutional} & 0.086 & 0.279 & 0.094 & 0.035 & 0.128 \\ 
 ShellNet~\cite{chavan2022simultaneous} & 0.185 & 0.523 & 0.035 & 0.013 & 0.047  \\
 CenterSnap~\cite{irshad2022centersnap} & 0.180  & 0.526 & 0.037 & 0.006 & 0.042 \\
 \ours{} (Ours) & \textbf{0.290} & \textbf{0.649} & \textbf{0.031}  & \textbf{0.005} & \textbf{0.036}  \\
\bottomrule
\end{tabular}
\end{adjustbox}
 \caption{Comparison of methods for the task of 3D scene completion on YCB-V~\cite{xiang2018posecnn} and HOPE~\cite{lin2021fusion}. Higher numbers for the IoU and F-score metrics, and lower numbers for the Chamfer Distances (CD) indicate better performance.}
    \label{tab:sota-comparison}
    \vspace{-7mm}
\end{table}



Table~\ref{tab:sota-comparison} shows quantitative evaluations on within-training-distribution YCB-V dataset \cite{xiang2018posecnn} and out-of-training-distribution HOPE dataset \cite{tyree2022hope}. On YCB-V dataset, \ours{} is able to outperform CON and ShellNet on all 3D scene reconstruction metrics. CON takes as input a sparse point-cloud of the scene. When major parts of the input point clouds are missing, as the common case for single-view RGB-D point clouds, CON fails to infer those regions. ShellNet is trained to predict back-side depth image for the detected object. We notice that with varying viewing directions, ShellNet backside depths are either too thin or too thick resulting in low performance. MaskRCNN's failure to detect objects also directly contributed to lower performance for ShellNet. CenterSnap inputs RGB-D image and predicts object shapes via a multi-step procedure allowing CenterSnap to learn strong shape and pose priors for objects within training distribution. This allowed CenterSnap to perform strongly on YCB-V objects as it was trained on them, but we noticed it struggles with cases of occluded objects. \ours{} which is trained without ground truth object pose or shape supervision is able to match or outperform the baselines in all metrics. Fig. \ref{fig:qualitative-baselines} shows a qualitative comparison with baselines.

On the out-of-distribution HOPE dataset, \ours{} is able to outperform all baselines by an even larger margin. This shows that our normal and occlusion boundary-based depth completion method generalizes well to unseen novel scenes. Figure~\ref{fig:qualitative-geometry} shows qualitative results on these datasets. 

\begin{figure}[t]
    \centering
    \includegraphics[width=1\columnwidth]{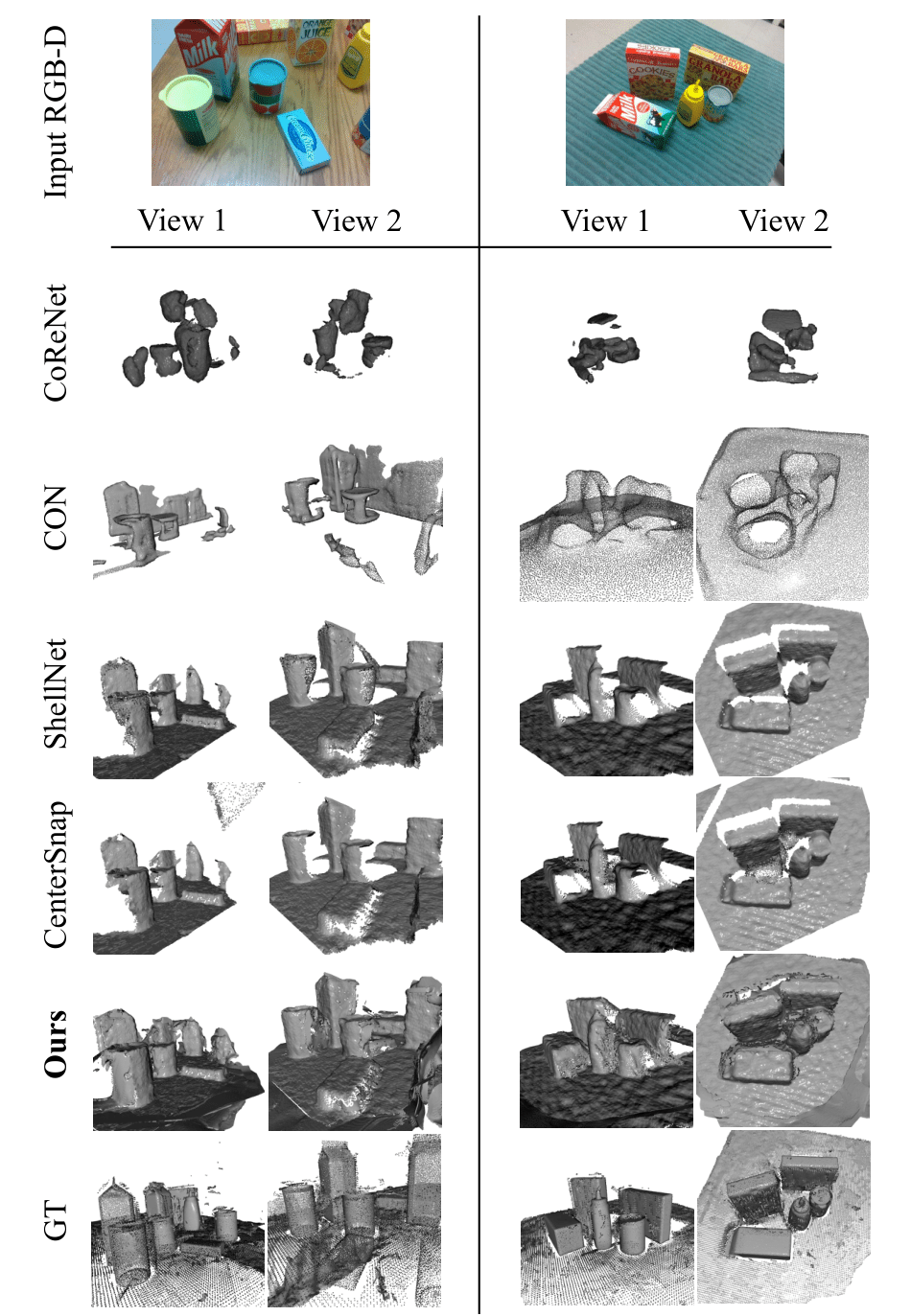}
    \caption{\textbf{Qualitative Comparison:} We compare our method against baselines for completing scene geometry given a single RGB-D image. Views 1 and 2 show novel viewpoints of the predicted point clouds from each method. Our method provides a denser and more complete reconstruction.}
    \label{fig:qualitative-baselines}
\end{figure}

As a byproduct, our method also produces novel views of unseen multi-object scenes from a single RGB-D image. Figure~\ref{fig:qualitative-inpainting} shows our method compared to the ground truth. We show that by combining our masking method with \dalle{}'s inpainting capability, realistic novels views can be generated for multiple unseen objects.


\subsection{Ablation Studies}

\begin{table}[ht!]
\centering
 \begin{tabular}{l | c c c} 
 \hline
 \textbf{Method} & IoU $\uparrow$ & F-Score $\uparrow$ & CD $\downarrow$ \\
 \hline
 \ours{} (S) & 0.262 & 0.613 & 0.038 \\
 \ours{} (G) & 0.261 & 0.613 & 0.038 \\
 \ours{} (Ours) & \textbf{0.290} & \textbf{0.649} & \textbf{0.036}  \\
 \hline
 \end{tabular}
 \caption{Prompt specificity  results  on  HOPE  dataset \cite{tyree2022hope}: \ours{} (S) denotes  our  model  with  scene  specific  prompt, \ours{} (G) uses ``household objects on a table" as the prompt for all scenes, and \ours{} (Ours) uses an image caption generator~\cite{li2022blip}.}
 \label{tab:prompt}
 \vspace{-3mm}
\end{table}

\textbf{Prompt Reliance:} Image diffusion models tend to heavily rely on the input prompt. To test our methods robustness, we performed an experiment using a general prompt (G), ``a photo of household objects on a table", for every scene to see how much performance degrades. We also use a specific prompt (S) where using the ground truth list of objects we list out every object on the table as the prompt. Table~\ref{tab:prompt} shows that our method does not largely depend on the type of prompt. We hypothesize that our view selection method retains enough surrounding context information in the input RGB image required for the inpainting model to inpaint successfully.

\textbf{Inpainting Model:}
We substitute in Stable Diffusion 2's~\cite{Rombach_2022_CVPR} inpainting model as an open-source alternative to \dalle{} in Table~\ref{tab:ablations}. We find that while accuracy decreases, it is still a viable inpainting substitute for our method.

\begin{figure}[t]
    \centering
    \includegraphics[width=\columnwidth]{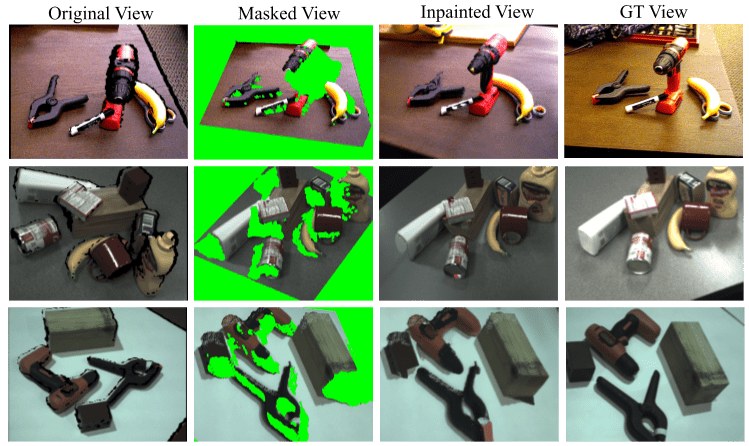}
    \caption{\textbf{Qualitative Novel View Results:} As a byproduct of our method we also show qualitative results for generating novel views of scenes from the the YCB-V dataset.}
    \label{fig:qualitative-inpainting}
\end{figure}


\textbf{Consistency Filtering:}
We test our method without applying the consistency filtering step by just combining all predicted viewpoints in Table~\ref{tab:ablations}. This caused a substantial decrease in accuracy as any hallucinated object is kept in.

\begin{table}[t]
\begin{adjustbox}{width=\columnwidth,center}
\centering
 \begin{tabular}{l | c c c c c} 
 \hline
 \textbf{Method} & IoU $\uparrow$ & F-Score $\uparrow$ & CD$(S^*, S)$ $\downarrow$ & CD$(S, S^*)$ $\downarrow$ & CD $\downarrow$ \\
 \hline
 \ours{} (SD-2) & 0.265 & 0.620 & 0.033 & \textbf{0.005} & 0.037 \\
  \ours{} (No Filter) & 0.271 & 0.574 & \textbf{0.017} & 0.030 & 0.047 \\ 
  \ours{} (Ours) & \textbf{0.290} & \textbf{0.649} & 0.031  & \textbf{0.005} & \textbf{0.036}  \\
 \hline
 \end{tabular}
  \end{adjustbox}
 \caption{Result of swapping out various parts of our method shown on the HOPE~\cite{lin2021fusion} dataset. (Ours) utilizes OpenAI's \dalle{} model and our consistency filtering method, (SD-2) uses Stable Diffusion 2's inpainting model, and (No Filter) refers to our method without filtering.}
\label{tab:ablations}
\vspace{-5mm}
\end{table}




\section{DISCUSSION}
We presented \ours{}, a novel method for 3D scene reconstruction. 
\ours{} solves the problem of 3D reconstruction of a cluttered scene of novel objects by leveraging the generalization capabilities of large visual language models. More specifically, our method utilizes the 2D inpainting capabilities of \dalle{} and generates a coherent set of inpainted views of the scene. It then lifts that information into 3D through a novel geometric multi-step method to finally output the point cloud of the reconstructed scene.

While we demonstrate the effectiveness of our method for generalizable scene completion, we also note that \dalle{} can generate unrealistic objects and parts of objects in the inpainted images. We mitigate this issue through various ways described in our method section, but these irregularities can adversely affect the reconstruction quality in a few cases.
While \ours{} shows the ability to complete the front and sides of objects in our scene, the backside of objects is often left incomplete due inpainting requiring enough context to accurately reconstruct the scene. At large angles away from the original viewpoint, the inpainting quality degrades due to the large areas of missing information - an exciting yet challenging problem for future work.





\balance
\bibliography{references}
\bibliographystyle{unsrt}

\clearpage
\section*{Appendix}

Here we include further details about implementation and experiments.

\subsection{Surface-Aware Masking Pseudocode}
We include pseudocode to help explain how our Surface-Aware Masking module (SAM) is implemented.

\begin{algorithm}
\caption{\textsc{Surface-Aware Masking (SAM)}}

\begin{algorithmic}
\Require{Input RGB-D image $\mathcal{I} = (\mathbf{I}, \mathbf{D})$, intrinsics $\mathbf{K}$, new viewpoint $\mathbf{T}_i$}
\State $U \leftarrow$ Subsample pixels from a uniform grid in $\mathcal{I}$
\State $X \leftarrow \{\}$ \Comment{initialize an empty point set.}
\ForAll {$\mathbf{u} \in U$}
\State $\mathbf{x} \leftarrow \mathbf{D}(\mathbf{u}) \mathbf{K}^{-1} \mathbf{u}$ \Comment{deprojection of $\mathbf{u}$ to 3D point $\mathbf{x}$.}
\For {$i \leftarrow 1$ to $m$}
    \State $\mathbf{p} \leftarrow \mathbf{x} + i \cdot c \cdot \mathbf{K}^{-1} \mathbf{u}$
    \State $X \leftarrow X \cup \{\mathbf{p}\}$ \Comment{set of points with equal spacing.}
\EndFor
\EndFor
\State $\mathcal{M} \leftarrow$ Mesh($X$) \Comment{surface triangulation to create a mesh.}
\State $\bar{\mathbf{I}}_i, \bar{\mathbf{D}}_i \leftarrow$ Reprojection of $\mathbf{I}, \mathbf{D}$ in camera viewpoint $\mathbf{T}_i$, where missing values are set to 0.
\State $\widetilde{\mathbf{D}}_i \leftarrow$ Depth map rendering of $\mathcal{M}$ in camera $\mathbf{T}_i$
\State $M \leftarrow \mathbf{0}_{H \times W}$ \Comment{initialize the mask image as zeros.}
\ForAll {$\mathbf{u} \in M$}
\State $M(\mathbf{u}) \leftarrow 1$ if $\bar{\mathbf{D}}_i(\mathbf{u}) = 0 \vee \bar{\mathbf{D}}_i(\mathbf{u}) > \widetilde{\mathbf{D}}_i(\mathbf{u})$ 
\EndFor

\State \textbf{return} $M, \bar{\mathbf{D}}_i$ 
\end{algorithmic}
\label{algo:surfawaremask}

\end{algorithm}

\subsection{Metrics}
We also include additional information about the metrics we use for quantitative results in our paper: 

\textbf{Intersection-over-Union (IoU)}:
We voxelize the ground truth and predicted point clouds at a fixed resolution and compute the IoU score by dividing the number of voxels that intersect to that of their union. In our experiments, we evaluate all the methods at the same grid resolution of $100^3$ after rescaling the predictions and ground truth to fit into the unit cube. \textbf{Chamfer Distance (CD)}: Chamfer distance is commonly used to measure the similarity between two point sets and is defined as:\vspace{-2mm}
\begin{equation}
    CD(X, Y) = \frac{1}{|X|} \sum_{\mathbf{x} \in X} \min_{\mathbf{y} \in Y} ||\mathbf{x} - \mathbf{y}||_2
\end{equation}
We separately report $CD(S, S^*)$ and $CD(S^*, S)$, as well as their their sum. $CD(S, S^*)$ measures how close the reconstructed points from $S$ are to the ground truth points $S^*$, whereas $CD(S^*, S)$ computes how well the ground truth shape is covered. \textbf{F-Score}: Following ~\cite{tatarchenko2019single}, we also report F-Score$@1\%$ which is a measure for the percentage of the surface points that were reconstructed correctly.

\subsection{Parameter Grid Search Experiment}
For choosing the number of viewpoints/viewpoint directions \(V\) as well as the context ratio \(C^*\) described in our method section, we perform a parameter search using 4 held out validation scenes from the YCB-V test set. We test using 6, 8, 10, and 12 viewpoints as well as a value of 0.3, 0.4, 0.5, 0.6, and 0.7 for our context threshold. We found that 10 views and 0.4 as a context threshold gave us the best accuracy on the validation set. 12 views and 0.4 as a context threshold performed similarly, but in the interest of runtime we use 10 for the final method. Table~\ref{tab:grid-search} shows the full results from this experiment.

\begin{table}[h!]
\vspace{-2mm}
\centering
\begin{tabular}{l|c c c c c}
\toprule
\(V\)~$\vert$~\(C^*\) & 0.3 & 0.4 & 0.5 & 0.6 & 0.7 \\
\midrule
6 & 0.064 & 0.053 & 0.051 & 0.057 & 0.064 \\
8 & 0.057 & 0.048 & 0.050 & 0.059 & 0.066 \\
10 & 0.053 & \textbf{0.047} & 0.052 & 0.059 & 0.070 \\
12 & 0.052 & \textbf{0.047} & 0.052 & 0.063 & 0.071 \\
\bottomrule
\end{tabular}
 \caption{Experiment using different values for context \(C^*\) and number of viewpoints \(V\) for our method on 4 validation scenes of the YCB-V~\cite{xiang2018posecnn} dataset using Chamfer Distance to indicate better performance.}
    \label{tab:grid-search}
    \vspace{-5mm}
\end{table}


\end{document}